\documentclass[runningheads,a4paper]{llncs}

\usepackage{amssymb}
\usepackage{graphicx}
\usepackage[utf8]{inputenc}
\usepackage{url}
\usepackage{tabularx}
\usepackage{epstopdf}

\urldef{\mailsa}\path|{pplonski, zaremba}@ire.pw.edu.pl|    
\newcommand{\keywords}[1]{\par\addvspace\baselineskip
\noindent\keywordname\enspace\ignorespaces#1}

\begin{document}

\mainmatter 
\title{Improving Performance of Self-Organising Maps with Distance Metric Learning Method.}
\titlerunning{Improving performance of Self-Organising Maps.}
\author{Piotr Płoński$^1$\and Krzysztof Zaremba$^1$}
\authorrunning{P. Płoński and K. Zaremba}
\institute{$^1$Institute of Radioelectronics, Warsaw University of Technology,\\ Nowowiejska 15/19,00-665 Warsaw, Poland,\\
\mailsa}
\maketitle
\footnotetext{The final publication is available at  \url{http://link.springer.com/chapter/10.1007/978-3-642-29347-4_20}}

\begin{abstract}

Self-Organising Maps (SOM) are Artificial Neural Networks used in Pattern Recognition tasks. Their major advantage over other architectures is human readability of a model. However, they often gain poorer accuracy. Mostly used metric in SOM is the Euclidean distance, which is not the best approach to some problems. In this paper, we study an impact of the metric change on the SOM's performance in classification problems. In order to change the metric of the SOM we applied a distance metric learning method, so-called 'Large Margin Nearest Neighbour'. It computes the Mahalanobis matrix, which assures small distance between nearest neighbour points from the same class and separation of points belonging to different classes by large margin. Results are presented on several real data sets, containing for example recognition of written digits, spoken letters or faces.

\end{abstract}

\section{Introduction}

Some real-world problems do not have an exact algorithmic solution. Currently,  there is a vast number of Artificial Intelligence(AI) methods which can be used to solve them. One of the branches of AI are Artificial Neural Networks. They are mathematical models inspired by biology. In 1982 T.Kohonen presented architecture called Self-Organising Maps (SOM) \cite{tuevo}, which provides a method of feature mapping from multi-dimensional space to usually a two-dimensional grid of neurons in  an unspervised way. This way of data analysis was proved as an  efficient tool in  many applications, both in academic and industrial solutions \cite{eng}. For example in character recognition tasks, image recognition tasks, face recognition \cite{aly}, analysis of words\cite{chinese}, grouping of documents \cite{klopotek}, visualisation \cite{duch}, and even bioinformatics (for example phylogenetic tree reconstruction \cite{dopazo}).

There exists a huge number of methods for improving SOM's performance. Some of them concentrated on finding an optimal size of a network\cite{growing}, faster learning \cite{faster} or applying different neighbourhood functions \cite{topology}. In this paper we investigate two additional improvements. The first one \cite{fessant}, \cite{aly} uses  Mahalanobis metric instead of the Euclidean one. The second improvement \cite{lasso} shows how to use SOM in a supervised manner.

In our approach, contrary to \cite{fessant}, \cite{aly}, \cite{bobrowski}, instead of computing the Mahalanobis matrix as an inverse of covariance matrix, it is  learned in a way assuring the smallest distance between points from the same class and large margin separation of points from different classes. Several algorithms exist for distance metric learning (DML) \cite{xing}, \cite{nca}. In this paper we use so-called Large Margin Nearest Neighbour (LMNN) method \cite{lmnn}. It introduces the distance metric learning problem as a convex optimization, which assures that the global minimum can be efficiently computed. First, we shortly describe SOM model used in supervised manner and LMNN method. Then we show how we combine these two approaches into our SOM+DML model. Finally we present results on real data sets.

\section{Methods}

Let's denote data set as $D = \{ ( \vec{x_i}, \vec{y_i})\}$, where $\vec{x_i}$ is an attribute vector of \textit{i}-th sample, and $\vec{y_i}$ is a class vector, where $\vec{y_{ij}}=0$ for $j \neq class_i$ and $\vec{y_{ij}}=1$ for $j = class_i$, where $class_i$ is class number for \textit{i}-th sample. 

\subsection{SOM model}

In this paper, we used the SOM architecture in a supervised manner so-called 'Learning Associations by Self-Organisation' (LASSO), first described in \cite{lasso}. The main difference between this SOM architecture and the original Kohonen's SOM architecture \cite{tuevo} is that during the learning phase the LASSO method takes into consideration class vector, additionally to attributes.  Herein, we used two-dimensional grid of neurons. Each neuron is represented by a weight vector $W_{pq}$, consisting of a vector corresponding to attributes $A_{pq}$, and to a class $C_{pq}$ ($W_{pq} = [A_{pq}; C_{pq}]$), where $(p,q)$ are indexes of the neuron in the grid. In a learning phase all samples are shown to the network in one epoch. For each sample we search for a neuron which is closest to the \textit{i}-th sample. The distance is computed by:
\begin{equation}
Dist_{train}(D_i, W_{pq}) = (\vec{x_i} - A_{pq})^T(\vec{x_i} - A_{pq}) + (\vec{y_i} - C_{pq})^T(\vec{y_i} - C_{pq}).
\end{equation}
The neuron $(p,q)$ with the smallest distance to \textit{i}-th sample is called the Best Matching Unit (BMU), we note its indexes as $(B_i, B_j)$. Once the BMU is found, the weight update step is executed. The weights of each neuron are updated with following formulas:
\begin{equation}\label{attr}
A_{pq} = A_{pq} + \eta (A_{pq} - \vec{x_i}),
\end{equation}
\begin{equation} 
C_{pq} = C_{pq} + \eta (C_{pq} - \vec{y_i}), 
\end{equation}
where $\eta$ is a learning coefficient, consisting of a learning step size parameter $\mu$ and a neighbourhood function $\tau$, so $\eta=\mu \tau$. Learning speed parameter is decreased between consecutive epochs, so that network's ability to remember patterns is improved. It is described by $\mu = \mu_{0} exp(- t \lambda)$, where $\mu_{0}$ is a starting value of the learning speed, $t$ is the current epoch and $\lambda$ is responsible for regulating the speed of the decrease. Neighbourhood function controls changing the  weights with respect to the distance to the BMU. It is noted as $\tau = exp(-\alpha S(B_i, B_j, p, q))$, where $\alpha$ describes the neighbourhood function width and $S(B_i, B_j, p,q)$ is the distance in the grid between the neuron and the BMU, computed by the following formula:
\begin{equation}
S(B_i, B_j, p, q) = (B_i - p)^2 + (B_j - q)^2.
\end{equation} 
We assumed a cost function as a sum of distances between samples and corresponding BMUs: 
\begin{equation}
F = \sum_l Dist_{train}(D_l, W_{B_i, B_j}).
\end{equation}
We train network till the cost function stops decreasing or a selected number of learning procedure iterations is exceed. 

The exploitation phase is performed after the learning phase. New samples, which do not take part in the training, are shown to the network in order to designate their class. It should be noted that only the part with attributes is presented to the network. The BMU is found by computing a distance between an attribute input vector and an attribute part of the weights using the following formula:
\begin{equation}
Dist_{test}(D_i, W_{pq}) = (\vec{x_i} - A_{pq})^T(\vec{x_i} - A_{pq}).
\end{equation}
For the tested sample, the designated class corresponds to position of maximum value in the part which code class information $C_{pq}$ in BMU weights.

\subsection{LMNN method}

In many cases the mostly used metric is an Euclidean one. It often gives poor accuracy, because it takes all dimensions with equal contribution and assumes no correlations between the  dimensions. Mahalanobis distance seems a better metric choice, because it is scale-invariant and takes into account input dimensions correlations. It is defined by:
\begin{equation}\label{mahal}
Dist_{M}(\vec{x_i},\vec{x_j}) = (\vec{x_i}-\vec{x_j})^T M(\vec{x_i}-\vec{x_j}), 
\end{equation}
where $M$ is usually an inverse of a covariance matrix. In case where $M$ is an identity matrix, the distance (\ref{mahal}) is equal to the Euclidean distance. 

In this paper, we learned Mahalanobis matrix using the method described in \cite{lmnn}. Matrix coefficients are computed in a way assuring large margin separation of points from different classes and a small distance between the points of the same class. Before starting the matrix learning for each point, $k$ nearest neighbours with the same class are found. They are called target neighbours, denoted by $\theta_{ij}=\{0,1\}$, where $\theta_{i,j}=1$ means that $x_j$ is the target neighbour of $x_i$, and  $\theta_{i,j}=0$ means otherwise. With no prior knowledge, the Euclidean distance can be used to point the target neighbours. Target neighbours are unchanged during the whole learning. Let's add a variable to indicate when the two samples have the same class, denoted as $y_{il}=\{0,1\}$, where $y_{il}=1$ when \textit{i}-th and \textit{l}-th samples are from the same class, zero when they are from different classes.

Finding an optimal matrix $M$ can be expressed as a semidefinite programming (SDP) optimization problem with the following cost function (\ref{cost}) and constrains (\ref{con1}), (\ref{con2}), (\ref{positive}):
\begin{equation}\label{cost}
minimize \sum_{ij} \theta_{i,j} Dist_M(\vec{x_i},\vec{x_j}) + c \sum_{i,j,l}\theta_{i,j}(1-y_{il})\xi_{ijl},
\end{equation} 
\begin{equation}\label{con1}
Dist_M(\vec{x_i},\vec{x_l}) - Dist_M(\vec{x_i},\vec{x_j}) \ge 1 - \xi_{ijl},
\end{equation}
\begin{equation}\label{con2}
\xi_{ijl} \ge 0,
\end{equation}
\begin{equation}\label{positive}
M \succeq 0.
\end{equation}
The first term in the minimization function penalizes a large distance between the samples and their target neighbours. The second term penalizes a small distance between the samples from different classes - it is expressed as slack variables $\xi_{ijl}$. The $c$ parameter balances the influence between these two terms. In this paper it is set to $0.5$, which gives equal strength to each term.  The constraint given in (\ref{positive}) requires the $M$ matrix to be positive semidefinite - all its eigenvalues should be nonnegative. Semidefinite programming is a convex optimization problem, so a global minimum can be efficiently computed. SPD can be solved using general purpose solvers, however \ in \ our \ approach \ we \ used \ Matalb implementation code\footnote{Matalb impementation of LMNN algorithm available from  \url{http://www.cse.wustl.edu/~kilian/code/code.html}} described in \cite{lmnn}, which is finely tuned to efficiently solve this kind of problems. Here most of slack variables are not used because samples are well separated. 

\subsection{SOM+DML model}

We are interested in a such linear transformation of sample attributes that will assure that the Euclidean distance computed on the transformed attributes will be equal to the Mahalanobis distance computed on the original attributes. Mahalanobis matrix $M$ can be written as $M=L^TL$, where $L$ is the searched transformation. Lets denote $\vec{u_i}$ as the transformed attributes of \textit{i}-th sample and $\vec{u_j}$ as the transformed attributes of \textit{j}-th sample:
\begin{equation}
\vec{u_i} = L \vec{x_i},
\end{equation}
\begin{equation}
\vec{u_j} = L \vec{x_j}.
\end{equation}
The distance between the transformed attributes in Euclidean distance should be equal to Mahalanobis distance between the original attributes:
\begin{equation}\label{MeqE}
Dist_{M}(\vec{x_i},\vec{x_j}) = Dist_{E}(\vec{u_i},\vec{u_j}).
\end{equation}
We will now search for $L$. Now for matrix $M$ we will find eigenvectors $\Phi$ and a matrix with eigenvalues on diagonal $\Lambda$:
\begin{equation}
M \Phi = \Phi \Lambda.
\end{equation}
Matrix $\Lambda$ can be expressed as:
\begin{equation}\label{eigenvalue}
\Phi^T M \Phi = \Lambda.
\end{equation}
Since matrix $M$ found by the LMNN algorithm is positive semidefine, diagonal elements in matrix $\Lambda$ are nonnegative. Thus, we can write: 
\begin{equation}\label{identity}
\Lambda^{- \frac{1}{2}} \Lambda \Lambda^{- \frac{1}{2}} = I.
\end{equation}
Substituting (\ref{eigenvalue}) into (\ref{identity}), we obtain:
\begin{equation} \label{m_i}
\Lambda^{- \frac{1}{2}} \Phi^T M \Phi \Lambda^{- \frac{1}{2}} = I.
\end{equation}
Now we can note $L$ as:
\begin{equation}\label{el}
L=\Lambda^{-\frac{1}{2}} \Phi^T.
\end{equation}
Using (\ref{el}) we can write (\ref{m_i}) as:
\begin{equation}
LML^T=I.
\end{equation}
We see that $M=L^TL$ and hence we can write:
\begin{equation}\label{ml}
(\vec{x_i}-\vec{x_j})^T M (\vec{x_i}-\vec{x_j}) = 
(L\vec{x_i}-L\vec{x_j})^T (L\vec{x_i}-L\vec{x_j})
\end{equation}
From (\ref{ml}) we see that (\ref{MeqE}) is true. This transformation of input attributes is also known as \textit{whitening transform}. It is worth mentioning that, in a transformation phase, only attributes $\vec{x}$ are transformed, the class part of input vector $\vec{y}$ is unchanged. Therefore, even though the data were pre-processed using the $L$ transformation, we still can use the original SOM algorithm. 

\section{Results}

Performance of the SOM+DML method was compared to the SOM model on six real data sets. As an accuracy measure we take the percentage of incorrect classifications. It is worth mentioning that getting the highest number of correct classifications is not the goal of this paper. Data sets are described in Table \ref{sets}. Sets 'Wine', 'Ionosphere', 'Iris', 'Isolet', 'Digits' are sets from the 'UCI Machine Learing Repository' \footnote{http://archive.ics.uci.edu/ml/}, set 'Faces' are from the 'The ORL Database of Faces'\footnote{http://www.cl.cam.ac.uk/research/dtg/attarchive/facedatabase.html}.

Now we briefly introduce the origin of the sets. Data sets 'Wine', 'Ionosphere', 'Iris' are classic benchmark sets, often used in testing newly developed classification algorithms. 'Isolet' data set represents a spoken letter recognition task. Its samples correspond to 26 letters of the alphabet. The original number of attributes (617) was projected by PCA to 100 principal components, which covers $93\%$ of its variance. 'Digits' data set represents a handwritten digit recognition problem. The samples are from 10 classes obtained from 43 people. Original images were 32x32 bitmaps downsampled to 64 attributes by creators. The face recognition task is presented by data set 'Faces'. It contains images of faces obtained from 40 people (40 classes). For each person 10 images were taken in different times, varying lightening, changing facial expressions and details. The original data - 92x112 pixels images in 256 gray levels was projected by PCA to 50 leading components ($83\%$ of variance), the so-called egienfaces method\cite{eigenfaces}. Each data set, if not originally divided to train/test subsets, was randomly divided by us - $70\%$ of data to training subset and $30\%$ to testing subset.

\begin{table}
\begin{center}
\begin{tabular}{c|  >{\centering}m{1.5cm} | >{\centering}m{1.5cm} | >{\centering}m{1.5cm} | >{\centering}m{1.5cm} | >{\centering}m{1.5cm} | c |}
\cline{2-7}
 & Wine & Ionosphere & Iris & Isolet & Digits & \ \ \ \ Faces \ \ \ \ \\ \hline 
\multicolumn{1}{|c|}{\ Train examples \ } & 126 & 246 & 105 & 6238   & 3823  & 280\\ \hline
\multicolumn{1}{|c|}{Test examples}  & 52  & 105 & 45  & 1559   & 1797  & 120\\ \hline
\multicolumn{1}{|c|}{Attributes}     & 13  & 34  & 4   & 100$^*$& 64    & 50$^*$ \\ \hline
\multicolumn{1}{|c|}{Classes}        & 3   & 2   & 3   & 26     & 10    & 40 \\ \hline
\multicolumn{1}{|c|}{Runs}           & 100 & 100 & 100 & 20     & 20    & 20 \\ \hline
\multicolumn{1}{|c|}{Net size}       & 4x4 & 6x6 & 6x6 & 20x20  & 20x20 & 20x20\\ \hline
\multicolumn{1}{|c|}{$k$ in LMNN}      & 2   & 5   & 1   & 4      & 3     & 3 \\ \hline
\end{tabular}
\end{center}
\caption{Description of data sets used to test performance of the LASSO+DML method and parameters of networks. The number of nearest neighbour in the LMNN was set by cross validation. ($^*$) In 'Isolet' and 'Faces' data sets, the number of attributes was reduced with PCA.}\label{sets}
\end{table}

\begin{figure}
\centering
\includegraphics[width=0.95\textwidth]{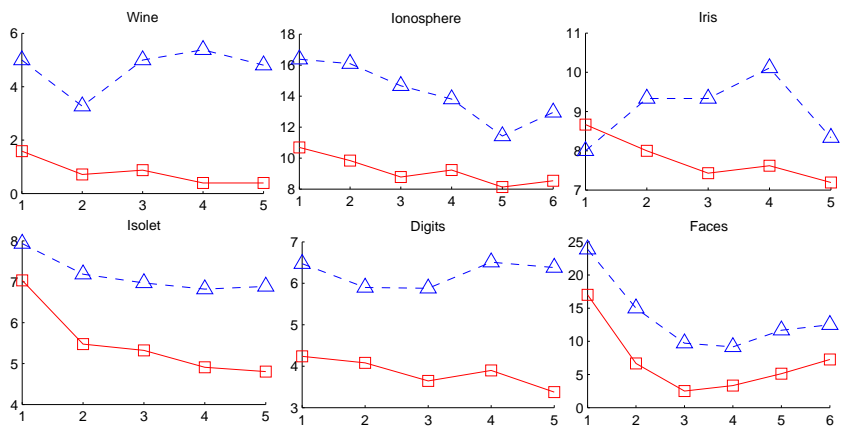}
\caption{Searching for optimal number of target neighbours in the LMNN method for all data set. On the x-axis there is $k$ number of target neighbours, on the y-axis is percent of incorrect classification. Triangles with dashed lines represent test error and squares with solid lines illustrate a train error.}\label{cross}
\end{figure}

For each data set, we arbitrarily chose the network size (selecting optimal network size is not in the scope of this paper). The network size for the SOM and the SOM+DML models was equal. For all data sets, the following values of the learning parameters were used: $\mu_0=0.01$, $\lambda=0.005$, $\alpha=0.1$. For each data set a number of $k$ target neighbours in the LMNN algorithm was selected using cross validation with ten times repetition. Fig.\ref{cross} presents results of a selecting the target neighbours parameter ($k$) for all sets. Resulting $k$ values are shown in Table \ref{sets}. SOM weights were initialized with random numbers drawn from a normal distribution with mean $0$ and standard deviation $0.5$. Several runs were performed for each data set (see Table \ref{sets}), so that local minimums were avoided. The final result is a mean over all runs. The SOM and the SOM+DML models were trained with identical number of iterations.

\begin{figure}
\centering
\includegraphics[width=0.95\textwidth]{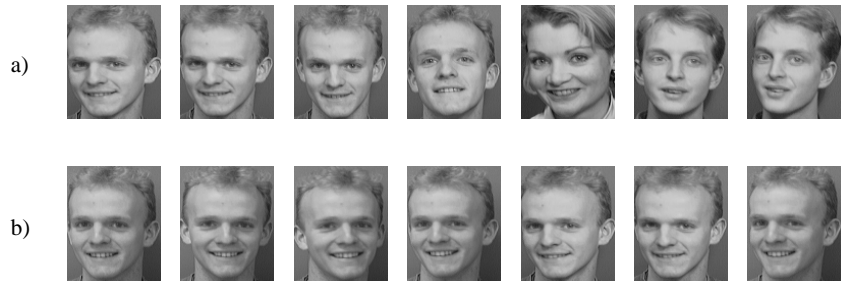}
\caption{Example of pictures from the 'Faces' data set classified as the same person by (a) the SOM model and by (b) the SOM+DML model.}\label{faces}
\end{figure}

The comparison of results obtained by the SOM and the SOM+DML method is presented in Table \ref{results}. With one exception the SOM+DML method achieves lower error rates in both training and testing subsets. On the 'Iris' data set the LMNN algorithm seems to cause the overfitting effect. It is clearly visible in Fig.\ref{cross} during search for the $k$ parameter. The greatest improvement was achieved on the 'Faces' set ($20.87\%$ over the SOM model). The comparison of example face pictures classified as belonging to the same class by the SOM method and the SOM+DML method is presented in the Fig.\ref{faces}. For the SOM+DML method, when using the 'Faces' set we observed a significant difference between the training error and the testing error. It is a small set, therefore the metric was well matched to the training set, giving small error. On 'Wine' and 'Isolet' data sets the improvement was $1.75\%$ and $1.89\%$ respectively. On the 'Digits' set there was a $2.68\%$ improvement on the testing subset, which corresponds to roughly $50$ digits. For the 'Ionosphere' data set the improvement was $6.67\%$. It is worth mentioning that for this set the largest $k$ value was used.

\begin{table}
\begin{center}
\begin{tabular}{c|c|c|c|c|c|c|c|c|c|c|c|c|}
\cline{2-13}
   & \multicolumn{2}{|c|}{Wine} & \multicolumn{2}{|c|}{Ionosphere} & \multicolumn{2}{|c|}{Iris} & \multicolumn{2}{|c|}{Isolet} & \multicolumn{2}{|c|}{Digits} & \multicolumn{2}{|c|}{Faces} \\ \cline{2-13}
   & Train & Test & Train & Test & Train & Test & Train & Test & Train & Test & Train & Test \\ \hline
   \multicolumn{1}{|c|}{SOM}     & 4.66 & 5.31  & 16.26 & 18.10 & 9.14 & \textbf{7.78} & 7.90 & 9.20 & 6.89 & 8.84 & 26.84 & 32.75 \\ \hline

   \multicolumn{1}{|c|}{SOM+DML} & \textbf{1.05} & \textbf{3.56} &  \textbf{8.13} & \textbf{11.43} & \textbf{8.76} & 8.44 & \textbf{5.32} & \textbf{7.31} & \textbf{3.86} & \textbf{6.16}  & \textbf{4.52}  &  \textbf{11.88} \\ \hline
\end{tabular}
\end{center}
\caption{Percent of incorrect classification on training and testing subsets for the SOM and the SOM+DML method. Results are means over all runs.}\label{results}
\end{table}

\section{Conclusions}

A method of improving performance of the Self-Organising Maps in classification tasks was described. Linear transformation of data was performed before SOM training phase. Matrix for the transformation has been obtained from the LMNN algorithm, which computes Mahalanobis matrix while assuring large margin separation between the points of different classes. We called our method SOM+DML. Testing of the method was demonstrated on several data sets, focused on recognition of: faces, handwritten digits and spoken letters. Test results confirm that the distance metric learning method improves the performance of the SOM network. Finding the optimal matrix for linear transformation plays a crucial role in obtaining improved results. Matlab implementation of the SOM+DML model is available from \url{ http://home.elka.pw.edu.pl/~pplonski/som_dml}.

\end{document}